\documentclass[10pt,twocolumn,letterpaper]{article}

\usepackage[final]{cvpr}      % To produce the REVIEW version
\usepackage{amsmath}
\definecolor{cvprblue}{rgb}{0.21,0.49,0.74}
\usepackage[pagebackref,breaklinks,colorlinks,allcolors=cvprblue]{hyperref}
\usepackage{multirow}
\usepackage{booktabs}
\usepackage{colortbl}
\usepackage{arydshln}
\usepackage{comment}
\def\paperID{*****} % *** Enter the Paper ID here
\def\confName{CVPR}
\def\confYear{2025}
\usepackage{bm}

\newcommand{\bmP}{{\bm P}}
\newcommand{\bmA}{{\bm A}}
\newcommand{\bmI}{{\bm I}}

\newcommand{\bmx}{{\bm x}}

\DeclareMathOperator*{\argmax}{argmax} % thin space, limits underneath in displays

\newcommand{\Ours}{\textsc{GraphGPT-o}\xspace}

\title{\Ours: Synergistic Multimodal Comprehension and Generation on Graphs}

\author{Yi Fang\thanks{Equal contribution.}\\
New York University Shanghai\\
{\tt\small yf2722@nyu.edu}
\and
Bowen Jin\footnotemark[1]\\
University of Illinois at Urbana-Champaign\\
{\tt\small bowenj4@illinois.edu }
\and
Jiacheng Shen\footnotemark[1]\\
New York University Shanghai\\
{\tt\small js12556@nyu.edu }
\and
Sirui Ding\\
University of California San Francisco\\
{\tt\small sirui.ding@ucsf.edu }
\and
Qiaoyu Tan\\
New York University Shanghai\\
{\tt\small qiaoyu.tan@nyu.edu }
\and
Jiawei Han\\
University of Illinois at Urbana-Champaign\\
{\tt\small hanj@illinois.edu  } \\
}

\begin{document}
\maketitle

\newcommand{\yfang}[1]{\textcolor{blue}{YF: #1}}

\begin{abstract}
The rapid development of Multimodal Large Language Models (MLLMs) has enabled the integration of multiple modalities, including texts and images, within the large language model (LLM) framework.
However, texts and images are usually interconnected, forming a multimodal attributed graph (MMAG).
It is underexplored how MLLMs can incorporate the relational information (\textit{i.e.}, graph structure) and semantic information (\textit{i.e.,} texts and images) on such graphs for multimodal comprehension and generation.
In this paper, we propose \Ours, which supports omni-multimodal understanding and creation on MMAGs.
We first comprehensively study linearization variants to transform semantic and structural information as input for MLLMs.
Then, we propose a hierarchical aligner that enables deep graph encoding, bridging the gap between MMAGs and MLLMs.
Finally, we explore the inference choices, adapting MLLM to interleaved text and image generation in graph scenarios. 
Extensive experiments on three datasets from different domains demonstrate the effectiveness of our proposed method. Datasets and codes will be open-sourced upon acceptance.
% However, the incorporation of graph-based modalities remains relatively unexplored. Graphs, which represent relationships between entities, possess unique structural and relational information that can enhance language models in understanding complex data. 
% In this paper, we address this gap by introducing a novel multimodal dataset tailored for graph data. 
% We propose innovative methods for graph prompt design, enabling seamless integration of graph data into LLMs. 
% Additionally, we present a specialized hierarchical aligner to further processing graph information as input for LLM architectures. 
% Our work lays the groundwork for incorporating graph data into multimodal models and demonstrates how such data can enhance model comprehension and reasoning capabilities.
\end{abstract}

\section{Introduction}
\label{sec:intro}

Multimodal Large Language Models (MLLMs) \cite{dong2024dreamllmsynergisticmultimodalcomprehension,liu2024visual,sun2024generative,li2023blip} have made significant progress in recent years, allowing the comprehension and generation of diverse data modalities including text and images. 
However, in real-world scenarios, there exists a pervasive \textit{graph-structured relationships} between texts and images.
Such graph-structured relationship can be described as ``\textit{Multimodal Attributed Graphs}'' (MMAGs)~\cite{peng2024learningmultimodalgraphssurvey,jin2024instructg2isynthesizingimagesmultimodal,zhu2024multimodalgraphbenchmark}, where nodes are associated with image and text information.
For example, the artwork graph~\cite{mao2019visual} is composed of nodes that include images (pictures) and text (titles), with edges representing shared genres and authorship. This structure uniquely represents each artwork in relation to thousands of others within the graph, providing a context that extends beyond simple language descriptions or image references.
While MLLMs have demonstrated outstanding comprehension and generation capability for text and image data, it is questionable how they could utilize the structural information on MMAGs.

In this context, we formulate the problem of \textit{multimodal content generation} on MMAGs which tasks MLLMs with producing both a textual description and an accompanying image for a new node based on the graph connectivity and node attributes.
This task focuses on generating text-image pairs for a node from MMAGs, reflecting a wide range of practical applications. 
For example, generating an image and a text for a product node linked to others through co-purchase edges in an e-commerce MMAG is equivalent to recommending~\cite{deldjoo2024reviewmodernrecommendersystems,liu2024multibehaviorgenerativerecommendation} potential future products to users.
Likewise, creating an image and a title for a virtual artwork node in the art MMAG is comparable to creating virtual artwork~\cite{huang2022drawartdreamdiverse,epstein2023art} that reflects the subtle styles of various artists and genres.

However, directly adopting MLLMs on MMAGs for multimodal content generation presents several challenges:
(1) \textit{Graph Size Explosion}: Although MMAGs provide substantial context for image and text generation, inputting the entire local subgraph structure to a model is impractical due to the exponential increase in size with additional hops, leading to excessively long context sequences.
(2) \textit{Non-Euclidean Nature}: Unlike texts or images, which follow linear structures, graphs are inherently non-Euclidean with complex topologies~\cite{bronstein2017geometric}, making them challenging to feed into MLLMs. 
(3) \textit{Hierarchical Modality Dependency}: At the node level, complementary information from associated text and image data enhances the semantic understanding of individual nodes. At the subgraph level, integrated features derived from node text/image semantics and local graph structure enable a more nuanced understanding of the subgraph's context for target node generation.
(4) \textit{Inference Dependency}: Due to the intrinsic interdependence between text and image features within a node, as well as the dual objectives of image and text generation, the order of inference across these modalities is critical.

To address these challenges, we introduce \Ours, a multimodal large language model tailored for comprehensive understanding and creation within MMAGs. Our approach features several key contributions:
(1) We develop a personalized PageRank-based graph sampling method to extract relevant subgraph information, effectively mitigating the \textit{Graph Size Explosion} issue.
(2) We investigate various design approaches for graph linearization, adapting its \textit{Non-Euclidean Nature} to fit a sequential MLLM processing paradigm.
(3) We construct a hierarchical graph aligner, incorporating a node-level modality fusion Q-Former and a graph structure Q-Former to capture \textit{Hierarchical Modality Dependency} within MMAGs.
(4) We explore different inference strategies, including sequential and parallel generation, to address \textit{Inference Dependency} across modalities in MMAGs.
With adaptive graph prompt designs and specialized alignment techniques, \Ours achieves effective comprehension and content generation in MMAGs, overcoming key challenges related to graph topology and multimodal attribute integration.

The primary contributions of this paper are as follows:
\begin{itemize}
\item \textbf{Problem Formulation and Benchmarking.} We formally define the task of multimodal content generation from MMAGs and introduce three real-world benchmark datasets across domains such as art and e-commerce to support this task.
\item \textbf{Proposed Methodology.} We introduce \Ours, a multimodal large language model designed to effectively encode graph structures for concurrent image and text generation.
\item \textbf{Experiments and Evaluation.} We perform comprehensive experiments and evaluations, demonstrating that \Ours achieves significant improvements over baseline models.
\end{itemize}

\section{Problem Formulation}\label{sec:problem-formulation}

\subsection{Multimodal Attributed Graphs}
\newtheorem{definition}{Definition}
\begin{definition}{(Multimodal Attributed Graphs (MMAGs))}
A multimodal attributed graph is defined as $\mathcal{G}=(\mathcal{V}, \mathcal{E}, \mathcal{P}, \mathcal{D})$, where $\mathcal{V}$, $\mathcal{E}$, $\mathcal{P}$, and $\mathcal{D}$ denote the sets of nodes, edges, images, and documents, respectively. Each node $v_i\in \mathcal{V}$ contains corresponding image information $p_{v_i}\in\mathcal{P}$ and textual information $d_{v_i}\in \mathcal{D}$.
\end{definition}
Some examples of MMAGs include (1) e-commerce produce graphs ($\mathcal{G}$), where product nodes ($v\in\mathcal{V}$) are associated with product image ($p \in \mathcal{P}$) and title ($d \in \mathcal{D}$); and (2) artwork graphs ($\mathcal{G}$), where artwork nodes ($v\in\mathcal{V}$) contain picture ($p \in \mathcal{P}$) and title ($d \in \mathcal{D}$).

\subsection{Problem Definition}\label{sec:prob-def}

\begin{definition}{(Node Multimodal Content Generation on MMAGs)}
In a multimodal attributed graph $\mathcal{G}=(\mathcal{V}, \mathcal{E}, \mathcal{P}, \mathcal{D})$, given a node $v_i\in\mathcal{V}$ within the graph $\mathcal{G}$, the goal is to generate $p_{v_i}$ and $d_{v_i}$, the corresponding image and text at $v_i$, with a learned model $(\hat{p}_{v_i}, \hat{d}_{v_i}) = f(v_i, \mathcal{G})$.
\end{definition}
This problem has numerous real-world applications.
In the context of e-commerce, this translates to generating an image ($p_{v_i}$) and a title ($d_{v_i}$) for a product ($v_i$) based on a user’s purchase history ($\mathcal{G}$), framing it as a generative recommendation task.
In the art domain, this involves generating an image ($p_{v_i}$) and a title ($d_{v_i}$) for an artwork ($v_i$) based on the associated artist's style or genre ($\mathcal{G}$), positioning it as a virtual artwork creation task.

\section{Methodology}
\label{sec:methodology}
In this section, we present our \Ours framework, a novel approach for generating image-text pairs on MMAGs using multimodal LLMs (MLLMs). 
We begin by introducing graph information into MLLMs in Section \ref{sec:graph-mllm}.
Next, we describe a personalized PageRank-based graph sampling strategy in Section~\ref{sec:sampling}, addressing the \textit{Graph Size Explosion} challenge.
In Section~\ref{sec:graph-tokenization}, we propose graph linearization strategies and develop a hierarchical graph aligner to address the \textit{Non-Euclidean Nature} of graphs and capture \textit{Hierarchical Modality Dependency} in MMAGs.
Finally, in Section~\ref{sec:inference-strategy}, we explore different generation strategies to manage \textit{Inference Dependency} across modalities.

% Next, we explore various generation strategies for the inference stage in Section~\ref{sec:inference-strategy}. 
% Finally, in Section~\ref{sec:graph-alignment}, we present the \Ours framework, which effectively aligns graph information with image-text pair information.

\begin{figure*}
    \centering
    \includegraphics[width=1\linewidth]{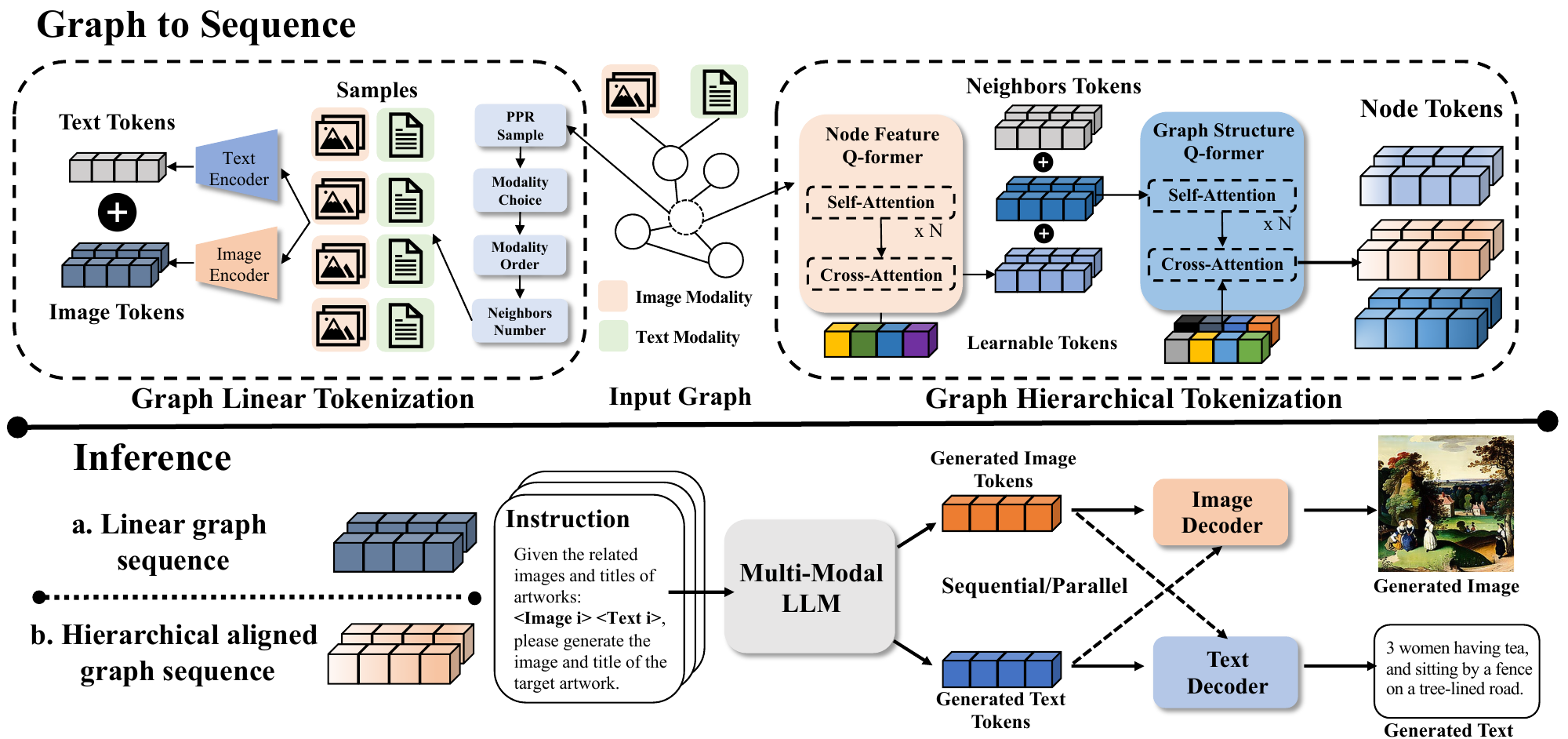}
    \caption{The overall framework of the proposed \Ours is as follows. Given a target node in a multimodal attribute graph (MMAG), we begin by using personalized PageRank for neighbor sampling. These sampled neighboring nodes are then fed into a Hierarchical Multimodal Aligner, which aligns text, image, and graph structure data. Each modality of a node is initially encoded and fused through multiple self-attention and cross-attention layers to produce multimodal node tokens. Subsequently, the tokens are processed by a graph structure Q-former, ultimately serving as inputs to the Multimodal LLM.}
    \label{fig:pipeline}
\end{figure*}

\subsection{Multimodal LLM on MMAGs}\label{sec:graph-mllm}
\textbf{DreamLLM.} The proposed \Ours is built upon DreamLLM~\cite{dong2024dreamllmsynergisticmultimodalcomprehension}, an MLLM capable of comprehension and generation on both text and image modalities.
To be specific, DreamLLM represents both texts and images as tokens and conducts encoding and generation in an autoregressive fashion:
\begin{equation}
\mathcal{L}^{\text{DreamLLM}}_{\text{MLLM}} = -\mathbb{E}_{t} \left[ \log p\left(x^{\text{WI}}_t \mid \bmx^{\text{WI}}_{<t} \right) \right],
\end{equation}
where $\bmx^{\text{WI}}=\{x^{\text{WI}}_t\}^T_{t=1}$ is an interleaved sequence containing both word tokens $\bm{w}=\{ w_i\}^N_{i=1}$ and image tokens $\bm{I}=\{I_k\}^K_{k=1}$.
The generated image tokens are decoded into images with a Stable Diffusion \cite{rombach2022high} trained by the following objective:
\begin{equation}
\mathcal{L}^{\text{DreamLLM}}_{\text{SD}} = \mathbb{E}_{t, \epsilon}\left[ \left| \epsilon - \epsilon_{\theta}\left(z_t; C^{\text{WI}}_{\text{DreamLLM}}, t \right) \right|^2 \right],
\end{equation}
where $C^{\text{WI}}_{\text{DreamLLM}}$ represents the condition incorporating information from both text and image modalities.

\vspace{0.1in}
\noindent \textbf{\Ours: introducing graph signals into MLLMs.}
In the context of MMAGs, generating image and text content for a node $v_i$ requires utilizing semantic information from the surrounding nodes within the \textit{graph} structure.
Therefore, we introduce an auxiliary set of graph tokens $\bm{g}_{v_i}=\{ g_j\}^M_{j=1}$ as input to the MLLM in addition to the text and image tokens:
\begin{equation}
\mathcal{L}^{\text{\Ours}}_{\text{MLLM}} = -\mathbb{E}_{t} \left[ \log p\left(x^{\text{WIG}}_t \mid \bmx^{\text{WIG}}_{<t} \right) \right],
\end{equation}
where $\bmx^{\text{WIG}}=\{x^{\text{WIG}}_t\}^T_{t=1}$ is an interleaved sequence containing both word tokens $\bm{w}$, image tokens $\bm{I}$ and graph tokens $\bm{g}$.
The image decoder Stable Diffusion \cite{rombach2022high} is then trained by the following objective:
\begin{equation}
\mathcal{L}^{\text{\Ours}}_{\text{SD}} = \mathbb{E}_{t, \epsilon}\left[ \left| \epsilon - \epsilon_{\theta}\left(z_t; C^{\text{WIG}}_{\text{\Ours}}, t \right) \right|^2 \right],
\end{equation}
where $C^{\text{WIG}}_{\text{\Ours}}$ represents the condition incorporating information from text, image, and graph modalities.

\subsection{Personalized PageRank Neighbor Sampling.}\label{sec:sampling}
A simple approach to obtain $\bm{g}_{v_i}$ is to encode the entire local subgraph of $v_i$ within $\mathcal{G}$.
However, this becomes impractical as the subgraph size grows exponentially with each additional hop, resulting in excessively long context sequences.
Additionally, irrelevant or extraneous information in the local subgraph could misguide the model.
To overcome this, inspired by \cite{gasteiger2022predictpropagategraphneural}, we utilize personalized PageRank (PPR) to selectively gather information for constructing $\bm{g}_{v_i}$ from a graph structure perspective.

To be specific, PPR \cite{haveliwala2002topic} utilizes the graph structure to produce a ranking score, $P_{i,j}$, for each node $v_j$ relative to a target node $v_i$. A higher score $P_{i,j}$ indicates a stronger “similarity” or relevance between nodes $v_i$ and $v_j$. We represent the PPR scores across all nodes with the PPR matrix $\bmP\in \mathbf{R}^{n\times n}$, where each row $P_{i,:}$ corresponds to the PPR vector for node $v_i$. The PPR matrix $\bmP$ is computed by solving the following equation:
\begin{gather}\label{eq:ppr}
    \bmP = \beta \hat{\bmA}\bmP + (1-\beta)\bmI.
\end{gather}
where $\beta$ is the reset probability governing the random walk in PPR and $\hat{\bmA}$ is the normalized adjacency matrix.
Using the PPR matrix, we define the PPR-based graph neighborhood nodes $N(v_i)$ to calculate $\bm{g}_{v_i}$ as the top $K$ most relevant neighbors, obtained by maximizing the sum of PPR scores for the selected neighbors:
\begin{gather}
   N(v_i) = \argmax_{N(v_i)\subset\mathcal{V}, |N(v_i)|=K} \sum_{v_j\in N(v_i)} P_{i, j}.
\end{gather}
This selection ensures that $N(v_i)$ captures the nodes most similar to $v_i$, based on their PPR scores.

\subsection{Multimodal Graph as Sequence}\label{sec:graph-tokenization}

After obtaining $N(v_i)$, the problem is how to extract meaningful graph representations from it. 
Given that \Ours takes sequential data as input, this involves tokenizing $N(v_i)$ into sequence $\bm{g}_{v_i}$. Previous studies ~\cite{zhu2024investigating, Liu2024graphprompter,instructglm} have explored methods for inputting text-attributed graphs as sequences into LLMs, but handling multimodal attributed graphs presents greater complexity.
In this section, we explore two ways to achieve this including (1) Linearization: simply linearizing the textual and image features in $N(v_i)$ into a sequence, and (2) Hierarchical Aligner: a hierarchical graph encoder to obtain deep representations as tokens for $N(v_i)$.

\subsubsection{Graph Linear Tokenization}

We first discuss tokenizing $N(v_i)$ with simple sequence linearization. This involves designing rules $\text{Linearize}(\cdot)$ to transform textual and image features in $N(v_i)$ into $\bm{g}_{v_i}$:
\begin{gather}
    \bm{g}_{v_i} = \text{Linearize}(N(v_i))
\end{gather}
Given that $N(v_i)$ is a set of nodes and each $v_j\in N(v_i)$ is associated with both text information $d_{v_j}$ and image information $p_{v_j}$, the design of the linearization rule should consider three factors: (1) modality choice; (2) modality order and (3) number of neighbors, which are discussed as follows:

% When constructing an instruction prompt for generation tasks using a sub-graph of nodes, each linked to an image-text pair, it is crucial to consider the \textbf{neighbor modality choices and order}, and \textbf{number of neighbors} in the prompt. These factors directly impact the prompt's clarity and effectiveness. We examine each below:

\vspace{0.05in}
\noindent
\textbf{Modality choice.} Depending on the graph, it is possible that presenting only texts $\{d_{v_j}| v_j\in N(v_i)\}$ or only images $\{d_{v_j}| v_j\in N(v_i)\}$ or both of them could benefit the multimodal content generation on MMAGs.
% For each node, deciding whether to include the image, text, or both is essential. While including both provides richer context, task-specific or computational constraints may favor using just one modality if it is more relevant to the task.

\vspace{0.05in}
\noindent
\textbf{Modality order.} Given that we have both text and image modality, it is flexible to adjust the order of different information, including (1) all images first, followed by texts, (2) all texts first, followed by images, and (3) interleaving image and text for each node $v_j \in N(v_i)$.
% The sequence of information is another key decision. Options include inputting all images first, followed by texts, or interleaving image and text for each node. The chosen order affects the model’s understanding and coherence of the prompt.

\vspace{0.05in}
\noindent
\textbf{Number of neighbors.} $N(v_i)$ is a list of nodes ranked by PPR score. Including more neighbors $v_j\in N(v_i)$ into $\bm{g}_{v_i}$ could potentially add more information but at the same time increase noise.
% Choosing the number of neighbors in the prompt affects context richness and length. While more neighbors offer broader context, too many can dilute focus. Selecting a few, highly relevant neighbors can optimize clarity and coherence.

In Section \ref{sec:res-linear}, we conduct systematic experiments on how different design choices affect the model performance.
After the design choice is given, $\{d_{v_j}| v_j\in N(v_i)\}$ are tokenized with text tokenizer and $\{d_{v_j}| v_j\in N(v_i)\}$ are tokenized with the pretrained CLIP encoder \cite{radford2021clip} similar to \cite{dong2024dreamllmsynergisticmultimodalcomprehension}. 

\subsubsection{Graph Hierarchical Tokenization}

Although linearization offers a solution for graph tokenization, it fails to capture hierarchical modality dependencies in MMAGs.
To be specific, at the node level, the combined information from associated text and image data contributes to a richer semantic representation of individual nodes. At the subgraph level, features synthesized from node-level semantics, alongside the local graph structure, enable a more comprehensive contextual understanding, thereby enhancing the generation of target nodes.
To this end, we design a hierarchical aligner $\mathcal{F}(\cdot)$ with a node feature Q-Former $\phi(\cdot)$ and a graph structure Q-Former $\psi(\cdot)$ to capture the node-level and subgraph-level modality dependency respectively:
\begin{gather}
    \bm{g}_{v_i} = \mathcal{F}(N(v_i)) = \psi(\{\phi(v_j) |v_j\in N(v_i)\})
\end{gather}

% \label{sec:graph-alignment}
% \subsubsection{Modality Fusion Q-Former}
\paragraph{Node Feature Q-Former.} It is proposed to learn node representations for $v_j\in N(v_i)$ considering the node-level modality dependency.
% To achieve alignment between text and image modalities, we introduce the Modality Fusion Q-Former, denoted as , which is designed to learn effective representations by integrating relevant text and image information for each neighboring node. 
As shown in Figure~\ref{fig:pipeline}, the Q-Former comprises two core Transformer \cite{vaswani2017attention} modules motivated by \cite{li2023blip}: (1) a self-attention module that facilitates deep information exchange between node text features and image features; (2) a cross-attention module that compresses node feature into a fixed number of representations.
% \begin{itemize}
% \item \textit{\textbf{Self-Attention}}: This module enables interactions among the hidden states within a single node, facilitating the capture of dependencies across text and image tokens.
% \item \textit{\textbf{Cross-Attention}}: This module applies attention to nodes in the neighborhood , weighted according to text guidance, thereby capturing dependencies between text data and neighboring image representations.
% \end{itemize}

% Let $\mathbf{H}^{(t)}_{v_j}\in \mathcal{R}^{d\times l}$ denote the hidden states calculated by the $t$-th Node Feature Q-Former layer.

The associated text $d_{v_j}$ and image $p_{v_j}$ of a node $v_j\in N(v_i)$ are first transformed into token representations $\bm{w}_{v_j}$ and $\bm{I}_{v_j}$ with text tokenizer and pretrained CLIP encoder respectively, which are then concatenated to form the initial input embedding:
\begin{equation}
\mathbf{H}^{(0)}_{v_j} = \left[ \bm{w}_{v_j} ;
\bm{I}_{v_j} \right] \in \mathbb{R}^{d \times (|d_{v_j}| + |p_{v_j}|)}
\end{equation}

% \noindent
% The text input is encoded into token embeddings, while the image input is also transformed into corresponding token embeddings. These text and image tokens are then concatenated to serve as input to the Q-Former. 

% \noindent
% For each node $\mathbf{{v_i}}$, the text data  is transformed into token embeddings:
% \begin{equation}
% \mathbf{X}_{d_{v_i}} = \left[ \bm{x}_1, \mathbf{x}_2, \ldots, \mathbf{x}_{|{d_{v_i}}|} \right] \in \mathbb{R}^{d \times |{d_{v_i}}|},
% \end{equation}
% where $\mathbf{x}_k$ represents the embedding for each token and $\mathbf{|d{v_i}}$ is the number of tokens in the text data.

% \noindent
% Similarly, each image  is encoded into token embeddings:
% \begin{equation}
% \mathbf{Z}_{p_{v_i}} = \left[ \mathbf{z}_1, \mathbf{z}_2, \ldots, \mathbf{z}_n \right] \in \mathbb{R}^{d \times n},
% \end{equation}
% where  $\mathbf{z}_k$ represents an image token embedding, and  is the number of tokens derived from the image data.

% \noindent
% The concatenated text and image tokens form the initial input embedding for node :
% \begin{equation}
% \mathbf{H}^{(0)}_{\mathcal{G}(v_i)} = \left[ \mathbf{X}_{d_{v_i}}
% \mathbf{Z}_{p_{v_i}} \right] \in \mathbb{R}^{d \times (|d_{v_i}| + n)}
% \end{equation}
% where  encapsulates both text and image information.

% \noindent
% Within each layer  of the Q-Former, a multi-head self-attention (MHA) mechanism is applied to the concatenated tokens to enable intra-node interactions, capturing dependencies across the text and image tokens:
The self-attention Transformer layers are designed to perform text and image modality information exchange calculated by:
\begin{equation}
\mathbf{H}^{(t)}_{v_j} = \text{Trans}_{\text{SAT}} \left( q, k, v=\mathbf{H}^{(t-1)}_{v_j} \right)
\end{equation}

% \noindent
Following $L_1$ self-attention Transformer layers, a cross-attention Transformer layer is applied, extracting the core feature into a fixed number of representations:
\begin{equation}
\mathbf{H}_{v_j} = \text{Trans}_{\text{CAT}} \left(q=\bm{Q}_V; k, v=\mathbf{H}^{(L_1)}_{v_j} \right)
\end{equation}
where $\bm{Q}_V\in \mathcal{R}^x$ is a node-level information aggregation soft prompt. The final representation $\mathbf{H}_{v_j}$ is leveraged as modality fused node feature representation.

% where the cross-attention process aggregates image embeddings from the neighborhood , utilizing text data to guide this integration.

% \noindent
% Finally, the multimodal representation  for each node is obtained as:
% \begin{equation}
% h_M(\mathcal{G}(v_i)) = \mathbf{H}^{(L)}_{\mathcal{G}(v_i)}
% \end{equation}
% where  represents the output of the Q-Former after  layers, capturing both intra-node text-image relationships and inter-node image dependencies to yield a comprehensive multimodal representation for each node.

% \subsubsection{Graph Structure Q-Former}
\paragraph{Graph Structure Q-Former.}
It is designed to aggregate the local context semantics inside $N(v_i)$, capturing the subgraph level modality dependency.
Similar to node feature Q-Former, graph structure Q-Former also contains two core Transformer modules:
(1) a self-attention module that enables deep information integration inside the local subgraph;
(2) a cross-attention module that aggregates the local semantics into a fixed number of representations.

% The Graph Structure Q-Former builds on the Modality Fusion Q-Former outputs, incorporating graph structure to aggregate multimodal representations from neighboring nodes. By leveraging graph topology, it enhances each node's representation by explicitly capturing structural relationships between nodes.

The node representations $\mathbf{H}_{v_j}$ for $v_j\in N(v_i)$ obtained from the node feature Q-Former are concatenated and serve as the initial inputs to the graph structure Q-Former:
\begin{equation} 
\mathbf{G}^{(0)}_{N(v_i)} = \left[{\mathbf{H}_{v_j} \ | \ v_j \in N(v_i)} \right] 
\end{equation} 

The self-attention Transformer layers are then applied to conduct deep information fusion between nodes inside the local subgraph:
\begin{equation}
\mathbf{G}^{(t)}_{N(v_i)} = \text{Trans}_{\text{SAT}} \left( q, k, v=\mathbf{G}^{(t-1)}_{N(v_i)} \right)
\end{equation}

% Following $L_2$ self-attention Transformer layers, a cross-attention Transformer layer is designed to 
After the $L_2$ self-attention Transformer layers, a cross-attention Transformer layer is designed to compress essential local graph features into a fixed set of representations:
\begin{equation}
\mathbf{G}_{N(v_i)} = \text{Trans}_{\text{CAT}} \left(q=\bm{Q}_G; k, v=\mathbf{G}^{(L_2)}_{N(v_i)} \right)
\end{equation}
where $\bm{Q}_G\in \mathcal{R}^x$ is a subgraph-level information aggregation soft prompt. 
The final representation is leveraged as graph token representations which are inputted into the MLLM: $\bm{g}_{v_i}=\mathbf{G}_{N(v_i)}$.

\subsection{Inference Strategy}\label{sec:inference-strategy}
Given the inherent interdependence between textual and visual features ($d_{v_i}$ and $p_{v_i}$) within a node $v_i$ and the joint objectives of generating both image and text, the order of inference across these modalities plays a crucial role.
To this end, we propose two strategies: (1) sequential inference and (2) parallel inference.

% In the inference phase, GraphGPT-o employs various strategies to generate multimodal content from prompts containing both text and images. These strategies enable effective interplay between modalities, enriching the generated output.

\vspace{0.05in}
\noindent
\textbf{Sequential Inference.} 
The proposed framework employs a sequential dual-generation process, in which one modality is generated first and subsequently serves as a conditioning factor for the generation of the other modality.
Specifically, this approach enables us to generate text  $d_{v_i}$ by optimizing $p(d_{v_i}|g_{v_i})$ and then generate the corresponding image $p_{v_i}$ by maximizing $p(p_{v_i}|g_{v_i}, d_{v_i})$.
Alternatively, we can initiate generation with the image $p_{v_i}$ by maximizing $p(p_{v_i}|g_{v_i})$ and then produce the text $d_{v_i}$ by optimizing $p(d_{v_i}|g_{v_i}, p_{v_i})$.
This sequential conditioning strategy ensures that the second generation step is contextually anchored in the outcome of the first, potentially enhancing coherence and consistency across modalities.

\noindent
\textbf{Parallel Inference.} 
The framework is designed to enable simultaneous dual generation of text and image by jointly optimizing $p(d_{v_i}|g_{v_i})$ and $p(p_{v_i}|g_{v_i})$.
This concurrent generation approach allows the production of $d_{v_i}$ and $p_{v_i}$ to proceed independently, mitigating the risk of error propagation from one modality serving as a conditional input for the other. 
Consequently, this parallel optimization strategy can reduce dependency on sequential conditioning, enhancing robustness in the generation process.

\section{Experiment}
\label{sec:experiment}

\subsection{Experimental Setups}
\noindent\textbf{Datasets.} We conduct experiments on three multimodal attributed graphs from distinct domains: ART500K, Amazon-Baby, and Amazon-Beauty. The ART500K dataset represents artworks, where nodes correspond to individual pieces, and edges indicate relationships such as shared authorship or genre. The Amazon datasets, comprising Amazon-Baby and Amazon-Beauty, represent product graphs. Here, nodes denote products, while edges capture co-view relationships. Each node in these graphs is enriched with a title and an image.

\noindent\textbf{Metrics.} To thoroughly assess the comprehension and generation capabilities of \Ours on multimodal attributed graphs, our evaluation focuses on two key aspects:

\begin{itemize} 
\item The quality of the synthesized image and text, and how well they align. 
\item The text/image correspondence between synthesized nodes and the conditioned sub-graphs.
\end{itemize}

To evaluate the quality of the synthesized outputs, we use CLIP (CLIP-I2) scores to compare the synthesized images with the ground truth images, assessing image generation quality. We also measure the perplexity of the generated text to evaluate its coherence. Additionally, we calculate the CLIP (CLIP-IT) score of generated image-text pairs to assess image-text alignment.

To evaluate alignment with the conditioned sub-graph, we calculate the KL divergence (KL-DV) between the distributions of the neighbor nodes and generated node image-text CLIP scores.

\subsection{Graph Linear Tokenization}
\label{sec:res-linear}
In this section, we study the quantitative results with graph linear tokenization, which are presented in Table~\ref{table-prompt}, from which we observe the following:
\newline \noindent \textbf{(1) Node Modality Integration.} Utilizing both modalities together generally improves model performance, indicating that integrating multiple information sources leads to a more comprehensive understanding of the data.
\newline \noindent \textbf{(2) Node Modality Order.} The order in which the modalities are processed does not consistently or significantly affect model performance.
\newline \noindent \textbf{(3) Inference Strategy.} Generating the image first typically enhances the quality of the synthesized image but may reduce text quality, whereas starting with text generation results in the lowest KL-DV score.

\begin{table*}[h!]
\centering
\resizebox{\textwidth}{!}{%
\begin{tabular}{c|c|c|cccc|cccc|cccc} 
\toprule
\multirow{2}{*}{\textbf{Modality}} & \multirow{2}{*}{\textbf{Order}} & \multirow{2}{*}{\textbf{Inference}} & \multicolumn{4}{c|}{\textbf{ART500K}} & \multicolumn{4}{c|}{\textbf{Beauty}} & \multicolumn{4}{c}{\textbf{Baby}} \\ 
\cline{4-15}
 &  &  & \textbf{CLIP-I2} & \textbf{Perplexity} & \textbf{CLIP-IT} & \textbf{KL-DV} & \textbf{CLIP-I2} & \textbf{Perplexity} & \textbf{CLIP-IT} & \textbf{KL-DV} & \textbf{CLIP-I2} & \textbf{Perplexity} & \textbf{CLIP-IT} & \textbf{KL-DV} \\ 
\hline\hline
\multirow{3}{*}{Text-only} & \multirow{3}{*}{Text-first} & Text-first & 65.83 & 163.3 & 22.66 & 4.65 & 55.49 & 193.8 & 24.6 & 10.72 & \textbf{78.89} & 328.3 & 17.88 & 2.51 \\
 &  & Image-first & 65.31 & 619.9 & \underline{24.54} & 5.04 & 65.56 & 668.5 & \textbf{25.64} & 14.72 & \underline{75.36} & 819.8 & 23.84 & 1.32 \\
 &  & Parallel & 65.31 & 158.5 & 16.37 & 9.96 & 65.56 & 206.9 & 19.99 & 22.31 & \underline{75.36} & 253.5 & 18.74 & 6.99 \\ 
\hline\hline
\multirow{3}{*}{Image-only} & \multirow{3}{*}{Image-first} & Text-first & 73.08 & 130.4& 19.53 & \underline{0.33} & 62.25 & \underline{124.5} & 9.49 & 18.55 & 72.40 & 155.5 & 27.8 & \textbf{0.73} \\
 &  & Image-first & 75.55 & 177.7 & 12.18 & 9.85 & \textbf{67.61} & 266.8 & 20.85 & 14.83 & 75.22 & \textbf{130.1} & 10.62 & 2.54 \\
 &  & Parallel & 75.55 & 460.8 & 22.66 & 5.76 & \textbf{67.61} & \textbf{108.8} & \underline{25.48} & 10.82 & 75.22 & 178.8 & 21.76 & 1.14 \\ 
\hline\hline
\multirow{9}{*}{Text-Image} & \multirow{3}{*}{Text-first} & Text-first & 71.15 & 555.7 & 18.86 & 0.49 & 60.5 & 514.7 & 20.84 & \textbf{9.85} & 67.98 & 402.3 & 27.59 & 1.93 \\
 &  & Image-first & \textbf{79.26} & \textbf{117.7} & 20.15 & 2.78 & \underline{66.89} & 379.7 & 6.78 & 10.64 & 59.27 & \underline{153.3} & 8.78 & 16.07 \\
 &  & Parallel & 79.26 & 737.7 & 22.56 & 3.82 & \underline{66.89} & 407.3 & 19.50 & 10.65 & 59.27 & 839.2 & 14.62 & 7.43 \\  
\cmidrule{2-15}
 & \multirow{3}{*}{Image-first} & Text-first & 74.14 & 217.3 & 23.81 & \textbf{0.19} & 62.19 & 259.3 & 22.83 & \underline{10.27} & 71.38 & 325.4 & \textbf{33.31} & 5.86 \\
 &  & Image-first & \underline{77.81} & 437.8 & 19.32 & 3.19 & 60.55 & 353.7 & 14.32 & 24.77 & 65.48 & 242.2 & 9.62 & 1.15 \\
 &  & Parallel & 77.81 & 219.8 & 22.18 & 2.97 & 60.55 & 207.1 & 22.51 & 22.21 & 65.48 & 169.9 & 23.42 & \underline{0.79} \\ 
\cmidrule{2-15}
 & \multirow{3}{*}{Interleaved} & Text-first & 68.40 & 315.7 & 18.57 & 0.70 & 64.70 & 310.8 & 25.5 & 10.39 & 71.71 & 522.5 & \underline{31.86} & \underline{0.79} \\
 &  & Image-first & 77.71 & \underline{117.9} & 20.84 & 2.66 & 64.64 & 346.5 & 6.79 & 10.63 & 62.62 & 572.5 & 21.98 & 0.88 \\
 &  & Parallel & 77.71 & 402.8 & \textbf{28.41} & 2.68 & 64.64 & 354.8 & 24.69 & 18.38 & 62.62 & 572.5 & 13.95 & 7.72 \\ 
\bottomrule
\end{tabular}}
\caption{Evaluation Results for Different Modalities and Orders on ART500K, Amazon-Beauty, and Amazon-Baby Datasets}
\label{table-prompt}
\end{table*}

\subsection{Graph Hierarchical Tokenization}
\label{sec:res-Hiera}
\subsubsection{Quantitative Evaluation.}
In this section, we compare the results of the original DreamLLM \cite{dong2024dreamllmsynergisticmultimodalcomprehension} and Chameleon \cite{team2024chameleon}, and DreamLLM fine-tuned with graph linear tokenization prompts named \Ours (Hard), and trained with an additional hierarchical aligner module named \Ours (soft). The default prompt setting for training and inference utilizes both modalities, with text-first in the instruction and generating text-first during inference. The results are shown in Table~\ref{table-main}, from which we can observe that \Ours(soft) outperforms baselines in most cases, especially aligns better with the golden sub-graph.

\begin{table*}[h!]
\centering
\resizebox{\textwidth}{!}{%
\begin{tabular}{c|cccc|cccc|cccc} 
\toprule
\multirow{2}{*}{\textbf{Model}} & \multicolumn{4}{c|}{\textbf{ART500K}} & \multicolumn{4}{c|}{\textbf{Beauty}} & \multicolumn{4}{c}{\textbf{Baby}} \\ 
\cline{2-13}
 & \textbf{CLIP-I2} & \textbf{Perplexity} & \textbf{CLIP-IT} & \textbf{KL-DV} & \textbf{CLIP-I2} & \textbf{Perplexity} & \textbf{CLIP-IT} & \textbf{KL-DV} & \textbf{CLIP-I2} & \textbf{Perplexity} & \textbf{CLIP-IT} & \textbf{KL-DV} \\ 
 \hline    
\cmidrule{1-13}
\multirow{1}{*}
{Janus} & 59.32 & 351.2 & 21.43 & 7.59& 42.52 & 415.6 & 17.8 & 1.45 & 52.81 & 324.5 & 25.6 & 2.43\\
\cmidrule{1-13}
\multirow{1}{*}
{Emu3} & 62.11 & 257.5 & 20.07 & 9.83 & 45.82 & 398.3 & 24.2 & 5.96 & 58.8 & 374.5 & 23.7 & 3.55 \\
\cmidrule{1-13}
\multirow{1}{*}{Chameleon} & 61.19 & 228.3 & 23.87 & 4.18 & 43.73 & 180.9 & 16.6 & 22.80 & 45.20 & 144.7 & 0.56 & 1.87 \\ 
\cmidrule{1-13}
\multirow{1}{*}{DreamLLM} & 71.15 & 555.7 & 18.86 & 0.4882 & 60.5 & 514.7 & 20.84 & 9.85 & 67.98 & 402.3 & 27.59 & 1.92 \\
\cmidrule{1-13}
\multirow{1}{*}{\Ours(Hard)} & \textbf{77.62} & 347.4 & 18.58 & 0.9377 & 57.99 & \textbf{107.9} & 24.66 & 12.75 & 68.23 & 124.8 & 20.24 & 1.39  \\
\cmidrule{1-13}
\multirow{1}{*}{ \Ours (Soft)} & 72.64 & \textbf{59.8} & \textbf{25.63} & \textbf{0.4327} & \textbf{63.46} & 285.0 & \textbf{27.38} & \textbf{5.82} & \textbf{74.77} & \textbf{103.2} & \textbf{31.14} & \textbf{0.23} \\
\bottomrule
\end{tabular}}
\caption{Results for different backbones on ART500K, Amazon-Beauty, and Amazon-Baby Datasets}
\label{table-main}
\end{table*}

\subsubsection{Qualitative Evaluation.}
We performed a qualitative evaluation by randomly selecting several generated cases, and comparing them with ground-truth, DreamLLM, and ChatGPT-4o. The results are presented in Figure 4, which includes sampled neighbor images and text from the graph alongside the ground truth images and text. These findings show that \Ours generates images that align closer with the contextual information derived from the golden sub-graph, while DreamLLM and ChatGPT- 4o stick to one style and fail to adapt based on the input. 

\begin{figure*}
    \centering
    \includegraphics[width=1\linewidth]{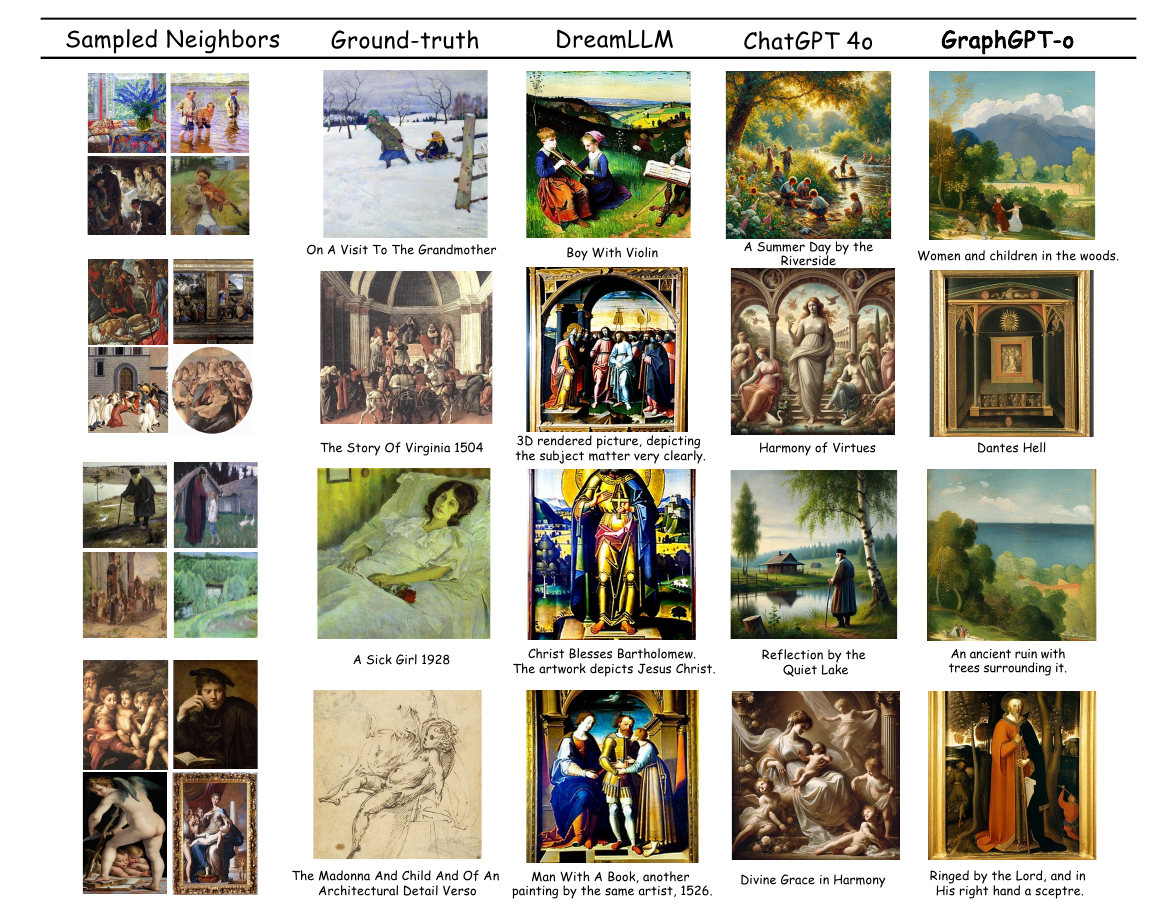}
    \caption{Qualitative evaluation. Our method exhibits better consistency with the ground truth by better utilizing the graph information from neighboring nodes.}
    \label{fig:enter-label}
\end{figure*}
\subsection{Ablation Study}
First, we evaluate the effectiveness of our \textbf{hierarchical aligner module} by individually removing the node feature Q-former and the graph structure Q-former. The results, presented in Table~\ref{table-ablation-module}, demonstrate that both modules contribute significantly to overall performance. Removing the graph structure causes a substantial increase in the KL-DV score while excluding the node features results in a higher perplexity for text.

\begin{table}
\centering
\resizebox{\linewidth}{!}{
    \begin{tabular}{c!{\vrule width \lightrulewidth}c!{\vrule width \lightrulewidth}c!{\vrule width \lightrulewidth}c!{\vrule width \lightrulewidth}c} 
    \toprule
    & CLIP-I2 & Perplexity & CLIP-IT & KL-DV \\ 
    \hline\hline
    \Ours w/o GSQ & 61.44 & \textbf{59.67} & \textbf{26.96} & 9.1406 \\ 
    \midrule
    \Ours w/o NFQ & {72.60} & 71.96 & {26.74} & 2.5050 \\ 
    \midrule
    \Ours & \textbf{72.64} & {59.80} & 25.63 & \textbf{0.4327} \\
    \bottomrule
    \end{tabular}
}
\caption{The impact of different modules in \Ours. w/o GSQ means without graph structure Q-Former and w/o NFQ means without node feature Q-Former.}
\label{table-ablation-module}
\end{table}

To further evaluate the effect of our hierarchical aligner module, a \textbf{GNN module} is used to replace it and the results are shown in Table ~\ref{table-ablation-GNN}, which shows that graph structure q-former is much better.

\begin{table}
\centering
\resizebox{\linewidth}{!}{
    \begin{tabular}{c!{\vrule width \lightrulewidth}c!{\vrule width \lightrulewidth}c!{\vrule width \lightrulewidth}c!{\vrule width \lightrulewidth}c} 
    \toprule
    & CLIP-I2 & Perplexity & CLIP-IT & KL-DV \\ 
    \hline\hline
    \Ours with GNN & 65.53   & 599.1      & 22.5    & 4.83 \\ 
    \midrule
    \Ours with GSQ & \textbf{71.15}   & \textbf{555.7}      & \textbf{18.86}   & \textbf{0.49} \\ 
    \bottomrule
    \end{tabular}
}
\caption{The impact of using GNN or graph structure Q-former (GSQ) for structure information learning in \Ours on ART500K dataset.}
\label{table-ablation-GNN}
\end{table}

\begin{figure*}
    \centering
    \includegraphics[width=0.7\linewidth]{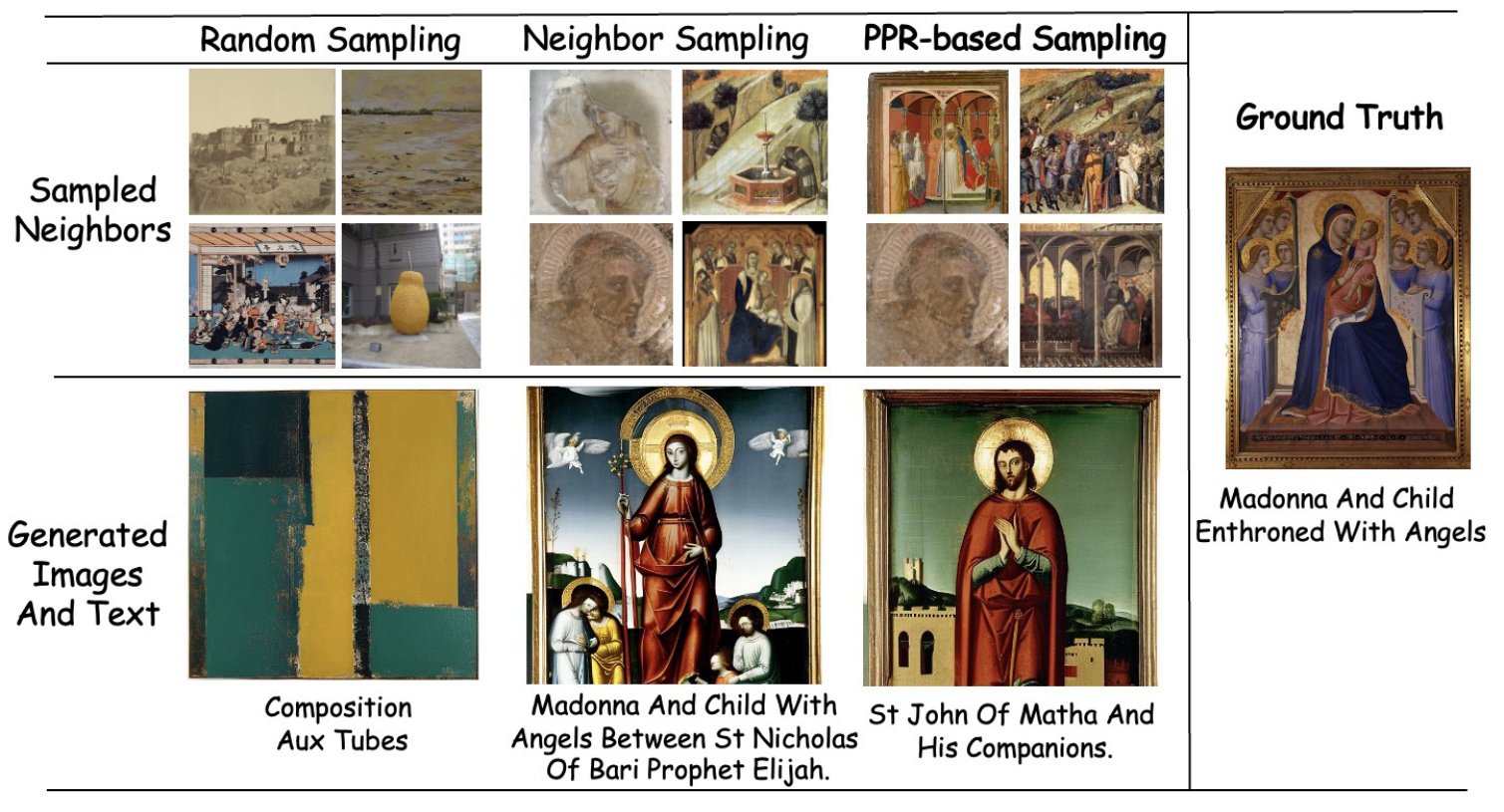}
    \caption{The impact of sampling strategies. Our proposed personalized PageRank sampling strategy leads to better image-text pair.}
    \label{fig-sample-strategy}
\end{figure*}

We then assess the impact of our \textbf{Personalized PageRank sampling} method. From Figure~\ref{fig-sample-strategy}, it can be observed that our proposed Personalized PageRank sampling strategy effectively captures neighbors that contribute most to the ground truth in terms of texture, artistic style, and visual consistency. This results in a generated image that more closely resembles the ground-truth image's detailed patterns and overall aesthetic.

\subsection{Other Studys}
\begin{figure}[h!]
    \centering
    \includegraphics[width=1.0\linewidth]{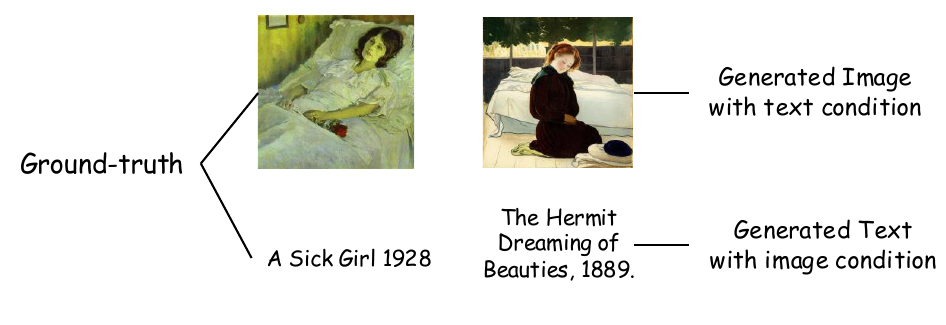}
    \caption{Study of \Ours generation with auxiliary node feature guidance: either image or text.}
    \label{fig-partial-guidance}
\end{figure}

\textbf{Study of generation with partial node feature guidance.} We further conduct a study on the performance of \Ours with additional node text guidance or node image guidance. From Figure~\ref{fig-partial-guidance} we can see that the style and the character information is well-captured.

\begin{figure}[h!]
\centering
\includegraphics[width=1.0\linewidth]{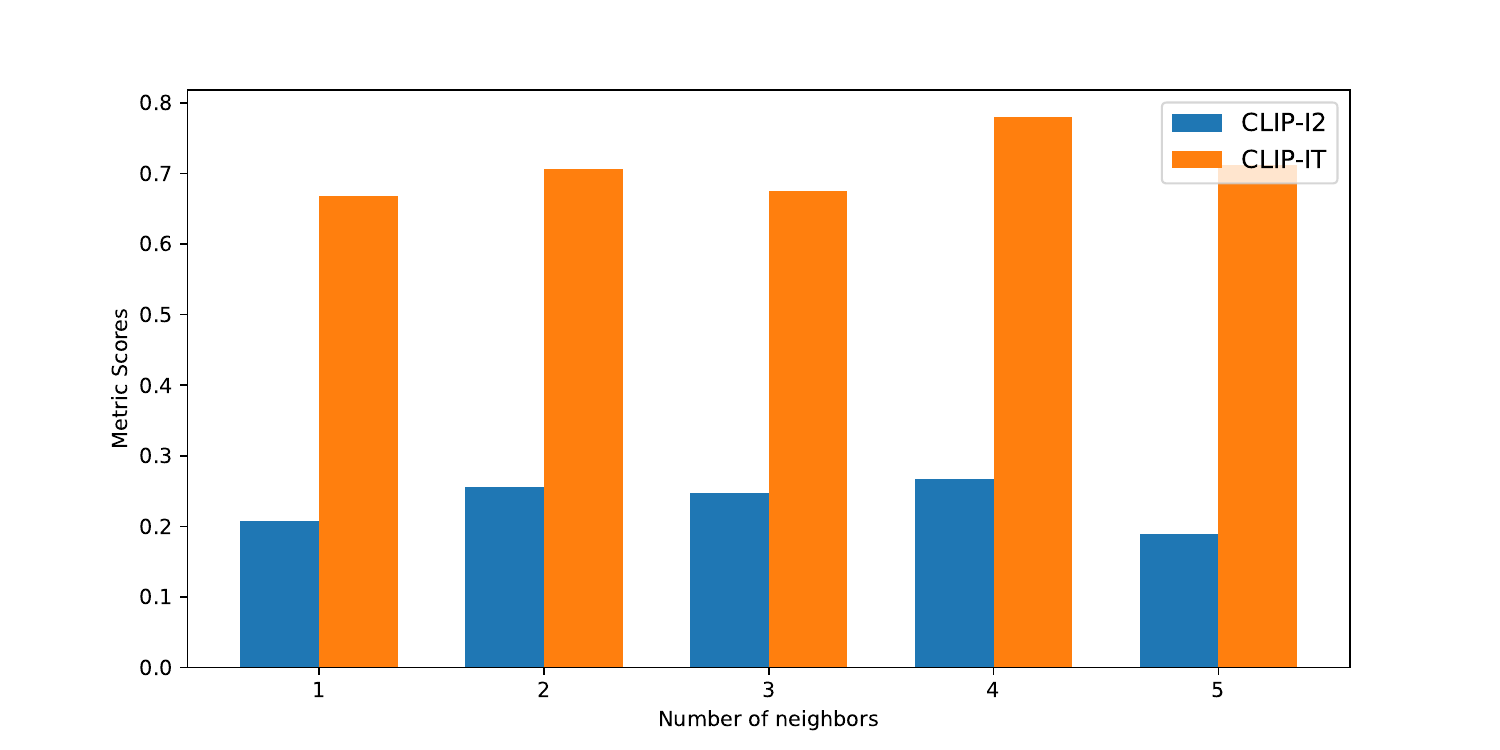}
    \caption{Study on the different number of neighbors on ART500K dataset.}
    \label{fig-neighbors-art}
\end{figure}

\noindent
\textbf{Study on the impact of number of neighbors.} Figure~\ref{fig-neighbors-art} shows that incorporating information from more neighbors can improve performance, but an excessive number may introduce noise, potentially hindering results.
\section{Related Work}\label{sec:relatedwork}

\subsection{Large Language Models on Graphs}
Large Language Models (LLMs) have driven substantial progress in graph learning applications \cite{jin2024large, ren2024survey}. Graph data, on the one hand, can be utilized directly to train LLMs \cite{chen2024graphwiz, zhang2023graph}. For instance, models like Heterformer \cite{jin2023heterformer} and Edgeformer \cite{ge2022edgeformer} introduce graph-enhanced Transformer architectures, positioning them as foundational models for graph-based LLMs. GraphGPT \cite{tang2024graphgpt} leverages graph structural data via graph instruction tuning, facilitating robust generalization across supervised and zero-shot graph learning tasks. LLaGA~\cite{chen2024llagalargelanguagegraph} employs a parameter-free GNN and incorporates the graph structure based on the order of node tokens. Similarly, InstructGraph \cite{wang2024instructgraph} employs a structured format verbalizer to encode graph data, enhancing LLMs in tasks requiring graph reasoning and generation. GraphAdapter \cite{huang2023can} incorporates GNNs as efficient adapters for LLMs, while GAugLLM \cite{fang2024gaugllm} advances self-supervised learning through augmented node features generated by an MoE module, effectively bridging textual and graph structures. UniGLM\cite{fang2024uniglmtrainingunifiedlanguage} uses structure information to build positive sample pairs in contrastive learning framework to train a unified text encoder. 
On the other hand, graph data can be utilized as external knowledge in a plug-and-play manner with LLMs \cite{jin2024graph, mavromatis2024gnn}. For example, Graph Chain-of-Thought \cite{jin2024graph} proposes an iterative framework that enables LLMs to reason, interact, and operate effectively on graphs. GNN-RAG \cite{mavromatis2024gnn} introduces a retrieval-augmented generation framework \cite{lewis2020retrieval}, using a GNN retriever to extract knowledge from graph data. Despite these advances, existing research has primarily focused on graphs with textual attributes, leaving multimodal attributed graphs underexplored.

% However, these approaches do not bridge the gap between MMAG and MLLM.

% In graph reasoning, LLMs address the challenge of interpreting complex relational structures by integrating with graph learning models or transforming graph data into language representations \cite{chen2024graphwiz,zhang2023graph}. 
% GraphLLM \cite{chai2023graphllm} enhances this by allowing LLMs to directly reason over graph data, moving beyond the Graph2Text approach. 
% Similarly, GraphText \cite{zhao2023graphtext} leverages LLMs' natural language capabilities to perform graph reasoning as text generation tasks by converting graph structures into syntax-based text sequences. 

% Graph learning with large language models (LLMs) has brought significant advancements in understanding complex relational data \cite{jin2024large,ren2024survey}. 
% Recent works integrating graph learning with LLM to build a general graph foundation model have emerged in multiple domains. 
% One for All (OFA)\cite{liu2023one} leverages text-attributed graphs and a graph prompting paradigm to generalize graph learning across graph tasks and domains, enabling in-context learning and robust performance without fine-tuning. 
% In node classification, LLMs enrich graph structure representations to better identify nodes in complex networks \cite{sun2023large,pan2024distilling}. 
% Graph LLM either improves graph learning tasks with LLM's text interpretation ability or improves LLM's reasoning ability with graph structural knowledge. 

\subsection{Multimodal Large Language Models}
Multimodal Large Language Models (MLLMs) have advanced the field by enabling unified multimodal understanding and generation within a single autoregressive framework \cite{yin2023surveymllm, zhang2024mm}. In terms of multimodal comprehension, models like Flamingo \cite{alayrac2022flamingo} process visual data interleaved with text, utilizing a gated cross-attention layer to encode inputs and produce free-form textual output. BLIP-2 \cite{li2023blip} introduces the Q-Former architecture, which maps images into a hidden space aligned with text tokens in LLMs, while LLaVA \cite{liu2024llava} simplifies this framework further with a projector and explores instruction tuning within the multimodal domain. Despite these advancements, current MLLMs primarily emphasize text generation and lack the capability to synthesize multimodal outputs (\textit{e.g.}, images).
To address this, DreamLLM \cite{dong2024dreamllmsynergisticmultimodalcomprehension} integrates an LLM backbone with a diffusion model to enable image generation as a multimodal output. Emu2 \cite{sun2024generative} scales this architecture to 37B parameters, demonstrating strong multimodal in-context learning and the ability to handle complex tasks requiring real-time reasoning, such as visual prompting and object-grounded generation. Chameleon \cite{team2024chameleon} proposes a stable training strategy from the ground up, featuring an alignment process and architectural parameterization tailored to early-fusion, token-based, mixed-modal settings. Nevertheless, most existing approaches overlook the relational dynamics between text and images, limiting their applicability to multimodal content generation tasks on multimodal attributed graphs (MMAGs).

\section{Conclusions}\label{sec:conclusions}

In this paper, we address the challenge of multimodal content generation on multimodal attributed graphs (MMAGs). 
To this end, we propose a graph-enhanced multimodal large language model, \Ours, designed with the following components:
(1) A personalized PageRank-based sampling strategy to extract informative neighbors from the graph, effectively mitigating the challenge of graph size explosion;
(2) A transformation mechanism that encodes graph information as sequences, employing either linearization or deep graph encoding with a hierarchical aligner, thereby addressing the non-Euclidean nature of graphs and hierarchical modality dependencies;
(3) Dual inference modes supporting both sequential and parallel inference to alleviate inference dependency issues.
We conduct comprehensive experiments on MMAGs within art and e-commerce domains, demonstrating the effectiveness of our approach against strong baseline methods. 
Future work includes extending MLLMs for discriminative tasks on MMAGs and capturing the complex heterogeneous relations between texts and images within these graphs.

{
    \small
    \bibliographystyle{ieeenat_fullname}
    \bibliography{main}

\begin{thebibliography}{41}
\providecommand{\natexlab}[1]{#1}
\providecommand{\url}[1]{\texttt{#1}}
\expandafter\ifx\csname urlstyle\endcsname\relax
  \providecommand{\doi}[1]{doi: #1}\else
  \providecommand{\doi}{doi: \begingroup \urlstyle{rm}\Url}\fi

\bibitem[Alayrac et~al.(2022)Alayrac, Donahue, Luc, Miech, Barr, Hasson, Lenc, Mensch, Millican, Reynolds, et~al.]{alayrac2022flamingo}
Jean-Baptiste Alayrac, Jeff Donahue, Pauline Luc, Antoine Miech, Iain Barr, Yana Hasson, Karel Lenc, Arthur Mensch, Katherine Millican, Malcolm Reynolds, et~al.
\newblock Flamingo: a visual language model for few-shot learning.
\newblock \emph{Advances in neural information processing systems}, 35:\penalty0 23716--23736, 2022.

\bibitem[Bronstein et~al.(2017)Bronstein, Bruna, LeCun, Szlam, and Vandergheynst]{bronstein2017geometric}
Michael~M. Bronstein, Joan Bruna, Yann LeCun, Arthur Szlam, and Pierre Vandergheynst.
\newblock Geometric deep learning: Going beyond euclidean data.
\newblock \emph{IEEE Signal Processing Magazine}, 34\penalty0 (4):\penalty0 18–42, 2017.

\bibitem[Chen et~al.(2024{\natexlab{a}})Chen, Li, Tang, and Li]{chen2024graphwiz}
Nuo Chen, Yuhan Li, Jianheng Tang, and Jia Li.
\newblock Graphwiz: An instruction-following language model for graph problems.
\newblock \emph{arXiv preprint arXiv:2402.16029}, 2024{\natexlab{a}}.

\bibitem[Chen et~al.(2024{\natexlab{b}})Chen, Zhao, Jaiswal, Shah, and Wang]{chen2024llagalargelanguagegraph}
Runjin Chen, Tong Zhao, Ajay Jaiswal, Neil Shah, and Zhangyang Wang.
\newblock Llaga: Large language and graph assistant, 2024{\natexlab{b}}.

\bibitem[Deldjoo et~al.(2024)Deldjoo, He, McAuley, Korikov, Sanner, Ramisa, Vidal, Sathiamoorthy, Kasirzadeh, and Milano]{deldjoo2024reviewmodernrecommendersystems}
Yashar Deldjoo, Zhankui He, Julian McAuley, Anton Korikov, Scott Sanner, Arnau Ramisa, René Vidal, Maheswaran Sathiamoorthy, Atoosa Kasirzadeh, and Silvia Milano.
\newblock A review of modern recommender systems using generative models (gen-recsys), 2024.

\bibitem[Dong et~al.(2024)Dong, Han, Peng, Qi, Ge, Yang, Zhao, Sun, Zhou, Wei, Kong, Zhang, Ma, and Yi]{dong2024dreamllmsynergisticmultimodalcomprehension}
Runpei Dong, Chunrui Han, Yuang Peng, Zekun Qi, Zheng Ge, Jinrong Yang, Liang Zhao, Jianjian Sun, Hongyu Zhou, Haoran Wei, Xiangwen Kong, Xiangyu Zhang, Kaisheng Ma, and Li Yi.
\newblock Dreamllm: Synergistic multimodal comprehension and creation, 2024.

\bibitem[Epstein et~al.(2023)Epstein, Kowalski, Thomas, and Zhang]{epstein2023art}
Ziv Epstein, John Kowalski, Laura Thomas, and Steve Zhang.
\newblock Art and the science of generative ai: A deeper dive.
\newblock \emph{arXiv preprint arXiv:2306.04141}, 2023.

\bibitem[Fang et~al.(2024{\natexlab{a}})Fang, Fan, Ding, Liu, and Tan]{fang2024uniglmtrainingunifiedlanguage}
Yi Fang, Dongzhe Fan, Sirui Ding, Ninghao Liu, and Qiaoyu Tan.
\newblock Uniglm: Training one unified language model for text-attributed graph embedding, 2024{\natexlab{a}}.

\bibitem[Fang et~al.(2024{\natexlab{b}})Fang, Fan, Zha, and Tan]{fang2024gaugllm}
Yi Fang, Dongzhe Fan, Daochen Zha, and Qiaoyu Tan.
\newblock Gaugllm: Improving graph contrastive learning for text-attributed graphs with large language models.
\newblock In \emph{Proceedings of the 30th ACM SIGKDD Conference on Knowledge Discovery and Data Mining}, pages 747--758, 2024{\natexlab{b}}.

\bibitem[Gasteiger et~al.(2022)Gasteiger, Bojchevski, and Günnemann]{gasteiger2022predictpropagategraphneural}
Johannes Gasteiger, Aleksandar Bojchevski, and Stephan Günnemann.
\newblock Predict then propagate: Graph neural networks meet personalized pagerank, 2022.

\bibitem[Ge et~al.(2022)Ge, Chen, and Wei]{ge2022edgeformer}
Tao Ge, Si-Qing Chen, and Furu Wei.
\newblock Edgeformer: A parameter-efficient transformer for on-device seq2seq generation.
\newblock \emph{arXiv preprint arXiv:2202.07959}, 2022.

\bibitem[Haveliwala()]{haveliwala2002topic}
Taher~H. Haveliwala.
\newblock Topic-sensitive pagerank.
\newblock In \emph{Proceedings of the 11th International Conference on World Wide Web (WWW '02)}.

\bibitem[Huang et~al.(2023)Huang, Zhang, Mei, and Ma]{huang2023can}
Jin Huang, Xingjian Zhang, Qiaozhu Mei, and Jiaqi Ma.
\newblock Can llms effectively leverage graph structural information: when and why.
\newblock \emph{arXiv preprint arXiv:2309.16595}, 2023.

\bibitem[Huang et~al.(2022)Huang, Tang, Dong, and Xu]{huang2022drawartdreamdiverse}
Nisha Huang, Fan Tang, Weiming Dong, and Changsheng Xu.
\newblock Draw your art dream: Diverse digital art synthesis with multimodal guided diffusion, 2022.

\bibitem[Jin et~al.(2023)Jin, Zhang, Zhu, and Han]{jin2023heterformer}
Bowen Jin, Yu Zhang, Qi Zhu, and Jiawei Han.
\newblock Heterformer: Transformer-based deep node representation learning on heterogeneous text-rich networks.
\newblock In \emph{Proceedings of the 29th ACM SIGKDD Conference on Knowledge Discovery and Data Mining}, pages 1020--1031, 2023.

\bibitem[Jin et~al.(2024{\natexlab{a}})Jin, Liu, Han, Jiang, Ji, and Han]{jin2024large}
Bowen Jin, Gang Liu, Chi Han, Meng Jiang, Heng Ji, and Jiawei Han.
\newblock Large language models on graphs: A comprehensive survey.
\newblock \emph{IEEE Transactions on Knowledge and Data Engineering}, 2024{\natexlab{a}}.

\bibitem[Jin et~al.(2024{\natexlab{b}})Jin, Pang, Guo, Wang, You, and Han]{jin2024instructg2isynthesizingimagesmultimodal}
Bowen Jin, Ziqi Pang, Bingjun Guo, Yu-Xiong Wang, Jiaxuan You, and Jiawei Han.
\newblock Instructg2i: Synthesizing images from multimodal attributed graphs, 2024{\natexlab{b}}.

\bibitem[Jin et~al.(2024{\natexlab{c}})Jin, Xie, Zhang, Roy, Zhang, Li, Li, Tang, Wang, Meng, et~al.]{jin2024graph}
Bowen Jin, Chulin Xie, Jiawei Zhang, Kashob~Kumar Roy, Yu Zhang, Zheng Li, Ruirui Li, Xianfeng Tang, Suhang Wang, Yu Meng, et~al.
\newblock Graph chain-of-thought: Augmenting large language models by reasoning on graphs.
\newblock \emph{arXiv preprint arXiv:2404.07103}, 2024{\natexlab{c}}.

\bibitem[Lewis et~al.(2020)Lewis, Perez, Piktus, Petroni, Karpukhin, Goyal, K{\"u}ttler, Lewis, Yih, Rockt{\"a}schel, et~al.]{lewis2020retrieval}
Patrick Lewis, Ethan Perez, Aleksandra Piktus, Fabio Petroni, Vladimir Karpukhin, Naman Goyal, Heinrich K{\"u}ttler, Mike Lewis, Wen-tau Yih, Tim Rockt{\"a}schel, et~al.
\newblock Retrieval-augmented generation for knowledge-intensive nlp tasks.
\newblock \emph{Advances in Neural Information Processing Systems}, 33:\penalty0 9459--9474, 2020.

\bibitem[Li et~al.(2023)Li, Li, Savarese, and Hoi]{li2023blip}
Junnan Li, Dongxu Li, Silvio Savarese, and Steven Hoi.
\newblock Blip-2: Bootstrapping language-image pre-training with frozen image encoders and large language models.
\newblock In \emph{International conference on machine learning}, pages 19730--19742. PMLR, 2023.

\bibitem[Liu et~al.(2024{\natexlab{a}})Liu, Li, Wu, and Lee]{liu2024llava}
Haotian Liu, Chunyuan Li, Qingyang Wu, and Yong~Jae Lee.
\newblock Visual instruction tuning.
\newblock \emph{Advances in neural information processing systems}, 36, 2024{\natexlab{a}}.

\bibitem[Liu et~al.(2024{\natexlab{b}})Liu, Li, Wu, and Lee]{liu2024visual}
Haotian Liu, Chunyuan Li, Qingyang Wu, and Yong~Jae Lee.
\newblock Visual instruction tuning.
\newblock \emph{Advances in neural information processing systems}, 36, 2024{\natexlab{b}}.

\bibitem[Liu et~al.()Liu, He, Tian, and Chawla]{Liu2024graphprompter}
Zheyuan Liu, Xiaoxin He, Yijun Tian, and Nitesh~V. Chawla.
\newblock Can we soft prompt llms for graph learning tasks?

\bibitem[Liu et~al.(2024{\natexlab{c}})Liu, Hou, and McAuley]{liu2024multibehaviorgenerativerecommendation}
Zihan Liu, Yupeng Hou, and Julian McAuley.
\newblock Multi-behavior generative recommendation, 2024{\natexlab{c}}.

\bibitem[Mao et~al.(2019)Mao, She, and Cheung]{mao2019visual}
Hui Mao, James She, and Ming Cheung.
\newblock Visual arts search on mobile devices.
\newblock \emph{ACM Transactions on Multimedia Computing, Communications, and Applications (TOMM)}, 15\penalty0 (2s):\penalty0 60, 2019.

\bibitem[Mavromatis and Karypis(2024)]{mavromatis2024gnn}
Costas Mavromatis and George Karypis.
\newblock Gnn-rag: Graph neural retrieval for large language model reasoning.
\newblock \emph{arXiv preprint arXiv:2405.20139}, 2024.

\bibitem[Peng et~al.(2024)Peng, He, and Xia]{peng2024learningmultimodalgraphssurvey}
Ciyuan Peng, Jiayuan He, and Feng Xia.
\newblock Learning on multimodal graphs: A survey, 2024.

\bibitem[Radford et~al.(2021)Radford, Kim, Hallacy, Ramesh, Goh, Agarwal, Sastry, Askell, Mishkin, Clark, et~al.]{radford2021clip}
Alec Radford, Jong~Wook Kim, Chris Hallacy, Aditya Ramesh, Gabriel Goh, Sandhini Agarwal, Girish Sastry, Amanda Askell, Pamela Mishkin, Jack Clark, et~al.
\newblock Learning transferable visual models from natural language supervision.
\newblock In \emph{International conference on machine learning}, pages 8748--8763. PMLR, 2021.

\bibitem[Ren et~al.(2024)Ren, Tang, Yin, Chawla, and Huang]{ren2024survey}
Xubin Ren, Jiabin Tang, Dawei Yin, Nitesh Chawla, and Chao Huang.
\newblock A survey of large language models for graphs.
\newblock In \emph{Proceedings of the 30th ACM SIGKDD Conference on Knowledge Discovery and Data Mining}, pages 6616--6626, 2024.

\bibitem[Rombach et~al.(2022)Rombach, Blattmann, Lorenz, Esser, and Ommer]{rombach2022high}
Robin Rombach, Andreas Blattmann, Dominik Lorenz, Patrick Esser, and Bj{\"o}rn Ommer.
\newblock High-resolution image synthesis with latent diffusion models.
\newblock In \emph{Proceedings of the IEEE/CVF conference on computer vision and pattern recognition}, pages 10684--10695, 2022.

\bibitem[Sun et~al.(2024)Sun, Cui, Zhang, Zhang, Yu, Wang, Rao, Liu, Huang, and Wang]{sun2024generative}
Quan Sun, Yufeng Cui, Xiaosong Zhang, Fan Zhang, Qiying Yu, Yueze Wang, Yongming Rao, Jingjing Liu, Tiejun Huang, and Xinlong Wang.
\newblock Generative multimodal models are in-context learners.
\newblock In \emph{Proceedings of the IEEE/CVF Conference on Computer Vision and Pattern Recognition}, pages 14398--14409, 2024.

\bibitem[Tang et~al.(2024)Tang, Yang, Wei, Shi, Su, Cheng, Yin, and Huang]{tang2024graphgpt}
Jiabin Tang, Yuhao Yang, Wei Wei, Lei Shi, Lixin Su, Suqi Cheng, Dawei Yin, and Chao Huang.
\newblock Graphgpt: Graph instruction tuning for large language models.
\newblock In \emph{Proceedings of the 47th International ACM SIGIR Conference on Research and Development in Information Retrieval}, pages 491--500, 2024.

\bibitem[Team(2024)]{team2024chameleon}
Chameleon Team.
\newblock Chameleon: Mixed-modal early-fusion foundation models.
\newblock \emph{arXiv preprint arXiv:2405.09818}, 2024.

\bibitem[Vaswani(2017)]{vaswani2017attention}
A Vaswani.
\newblock Attention is all you need.
\newblock \emph{Advances in Neural Information Processing Systems}, 2017.

\bibitem[Wang et~al.(2024)Wang, Wu, Hou, Liu, Gao, and McAuley]{wang2024instructgraph}
Jianing Wang, Junda Wu, Yupeng Hou, Yao Liu, Ming Gao, and Julian McAuley.
\newblock Instructgraph: Boosting large language models via graph-centric instruction tuning and preference alignment.
\newblock \emph{arXiv preprint arXiv:2402.08785}, 2024.

\bibitem[Ye et~al.(2024)Ye, Zhang, Wang, Xu, and Zhang]{instructglm}
Ruosong Ye, Caiqi Zhang, Runhui Wang, Shuyuan Xu, and Yongfeng Zhang.
\newblock Language is all a graph needs, 2024.

\bibitem[Yin et~al.(2023)Yin, Fu, Zhao, Li, Sun, Xu, and Chen]{yin2023surveymllm}
Shukang Yin, Chaoyou Fu, Sirui Zhao, Ke Li, Xing Sun, Tong Xu, and Enhong Chen.
\newblock A survey on multimodal large language models.
\newblock \emph{arXiv preprint arXiv:2306.13549}, 2023.

\bibitem[Zhang et~al.(2024)Zhang, Yu, Dong, Li, Su, Chu, and Yu]{zhang2024mm}
Duzhen Zhang, Yahan Yu, Jiahua Dong, Chenxing Li, Dan Su, Chenhui Chu, and Dong Yu.
\newblock Mm-llms: Recent advances in multimodal large language models.
\newblock \emph{arXiv preprint arXiv:2401.13601}, 2024.

\bibitem[Zhang(2023)]{zhang2023graph}
Jiawei Zhang.
\newblock Graph-toolformer: To empower llms with graph reasoning ability via prompt augmented by chatgpt.
\newblock \emph{arXiv preprint arXiv:2304.11116}, 2023.

\bibitem[Zhu et~al.(2024{\natexlab{a}})Zhu, Zhou, Qian, He, Zhao, Shah, and Koutra]{zhu2024multimodalgraphbenchmark}
Jing Zhu, Yuhang Zhou, Shengyi Qian, Zhongmou He, Tong Zhao, Neil Shah, and Danai Koutra.
\newblock Multimodal graph benchmark, 2024{\natexlab{a}}.

\bibitem[Zhu et~al.(2024{\natexlab{b}})Zhu, Huang, Jin, Jiao, Zhong, Chang, Lin, and Han]{zhu2024investigating}
Kerui Zhu, Bo-Wei Huang, Bowen Jin, Yizhu Jiao, Ming Zhong, Kevin Chang, Shou-De Lin, and Jiawei Han.
\newblock Investigating instruction tuning large language models on graphs.
\newblock \emph{arXiv preprint arXiv:2408.05457}, 2024{\natexlab{b}}.

\end{thebibliography}
}

\clearpage
\setcounter{page}{1}
\maketitlesupplementary

\section{Limitations}
In our current approach, we treat the graph as homogeneous, simplifying all nodes and edges into a single type. However, real-world graphs often consist of multiple node and edge types, each with unique semantic meanings. Future research could address this limitation by extending GraphGPT-o to heterogeneous graphs, allowing for richer and more nuanced representations of complex structures.

\section{Ethical Considerations}
GraphGPT-o presents a new method for improving the structural understanding of MLLMs through graph-based alignment. This approach seeks to tackle current issues in MLLMs, such as the uncontrolled generation of unsuitable content and susceptibility to adversarial attacks. Although GraphGPT-o provides enhancements, it still depends on the MLLM foundation, making it subject to these inherent limitations. Ethical concerns, like the potential for misuse, unintended generation of inappropriate content, and exposure to adversarial manipulation, need careful attention when deploying GraphGPT-o in practical applications.

\section{Experiment settings.}
For training, we randomly sampled 40,000 nodes from each original dataset. For testing, we randomly selected 50 nodes and its related neighbors from the rest of the dataset.

In the implementation of GraphGPT-o, we utilize DreamLLM as the pre-trained backbone. Within the Graph Hierarchical Tokenization module, the learnable tokens, as well as all self-attention and cross-attention layers, are randomly initialized. We employ a pre-trained CLIP encoder as the fixed image and text encoder, with an additional MLP to resolve dimensional discrepancies.

\begin{table}[h!]
\centering
\caption{Hyper-parameter configuration for model training.}
\begin{tabular}{|l|c|c|c|}
\hline
\textbf{Parameter}             & \textbf{ART500K} & \textbf{Beauty} & \textbf{Baby} \\ \hline
learning rate                      & 1e-5            & 1e-5           & 1e-5              \\ \hline
Batch size per GPU             & 1               & 1              & 1                 \\ \hline
warmup ratio          & 3e-3                & 3e-3              & 3e-3                  \\ \hline
Epochs                         & 1               & 1              & 1                 \\ \hline
loss weight of lm                     & 1              & 1             & 1                \\ \hline
loss weight of vm                  & 5             & 5            & 5               \\ \hline
\end{tabular}
\label{tab:hyperparams}
\end{table}

\section{More Experiment Results.}
We demonstrate more cases generated by DreamLLM and \Ours with comparision with the ground truth.

\begin{figure*}
    \centering
    \includegraphics[width=1\linewidth]{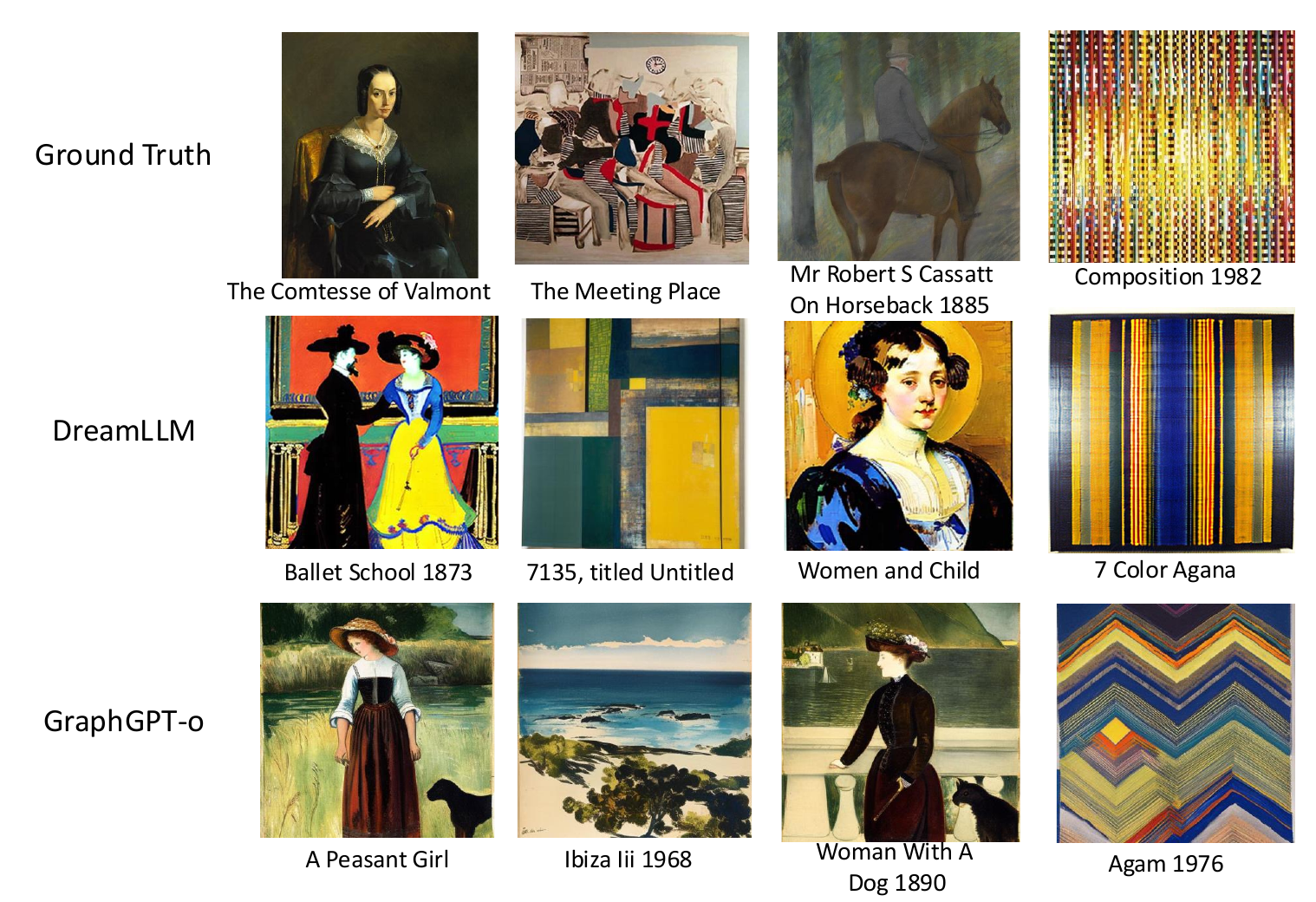}
    \caption{More cases for qualitative evaluation. Our method exhibits better consistency with the ground truth by better utilizing the graph information
from neighboring nodes}
    \label{fig:enter-label}
\end{figure*}

\end{document}